\documentclass[letterpaper, 10 pt, conference]{ieeeconf}  

\IEEEoverridecommandlockouts                              

\usepackage{cite}
\usepackage{amsmath,amssymb,amsfonts}
\usepackage{algorithmic}
\usepackage{graphicx}
\usepackage{textcomp}
\def\BibTeX{{\rm B\kern-.05em{\sc i\kern-.025em b}\kern-.08em
    T\kern-.1667em\lower.7ex\hbox{E}\kern-.125emX}}
\usepackage{dsfont}
\usepackage{paralist}
\usepackage{cite}
\usepackage{tabularx}
\usepackage{cancel}
\usepackage{bm}
\usepackage{url}
\usepackage{mathtools}
\usepackage{textcomp}
\usepackage{nccmath}
\usepackage{xcolor}
\usepackage{multirow}
\usepackage{hhline}
\usepackage{booktabs}
\usepackage{flushend}
\usepackage{hyperref}
\usepackage{microtype}
\usepackage[table]{xcolor}
\usepackage[hang,flushmargin]{footmisc}

\title{\LARGE \bf
Relationally Grounded Latent World Models for Autonomous Driving
}

\author{Fabian Schmidt$^{1,2}$, Markus Enzweiler$^{1}$, Abhinav Valada$^{2}$
\thanks{$^{1}$ Institute for Intelligent Systems, Esslingen University of Applied Sciences, Germany.}%
\thanks{$^{2}$ Department of Computer Science, University of Freiburg, Germany.}%
}

\begin{document}

\maketitle
\thispagestyle{empty}
\pagestyle{empty}

\begin{abstract}
Latent world models learn predictive representations for autonomous driving, but the relational semantics these states preserve often remain implicit.
We investigate whether traffic scene graphs can serve as \emph{privileged semantic supervision} for latent world representations.
Building on LAW, we construct actor-centric scene graphs from nuScenes 3D annotations, encode their serialized relational structure using a frozen text embedding model, and align the visual latent representations with this semantic target during training.
We remove the supervision branch at inference, so it requires neither scene graphs nor 3D annotations and adds no test-time computation.
On nuScenes, our method reduces average trajectory L2 error from $0.661$ to $0.622$ ($5.9\%$) and collision rate from $0.456$ to $0.217$ ($52.4\%$) relative to our retrained LAW baseline.
It also outperforms an unstructured caption-style semantic target, supporting the benefit of explicit relational structure for latent world-model representation learning.
\end{abstract}

\section{Introduction}
\label{sec:introduction}

End-to-end autonomous driving has evolved from unified perception--planning architectures toward foundation model-based agents with increasingly rich semantic representations.
Methods such as UniAD~\cite{hu2023planning} and VADv2~\cite{chen2024vadv2} jointly learn perception and planning representations, while language-aware approaches such as LMDrive~\cite{shao2024lmdrive}, DriveLM~\cite{sima2024drivelm}, SimLingo~\cite{renz2025simlingo}, and OpenDriveVLA~\cite{zhou2026opendrivevla} increasingly couple semantic reasoning with driving actions.
LAD-Drive~\cite{schmidt2026laddrive} further bridges the gap between high-level semantic intent and continuous planning through an action-aware diffusion decoder.
Together, these works highlight the value of semantically structured representations for driving while generally retaining semantic reasoning in the deployed model.

A complementary line of research investigates world models that learn predictive representations of driving scenes.
Generative driving world models predict future visual observations~\cite{hu2023gaia,wang2024drivedreamer,gao2024vista,mousakhan2025orbis}, whereas latent world models reason in compact feature spaces that integrate efficiently with downstream planning.
LAW~\cite{li2025enhancing} predicts future visual latents conditioned on the current scene representation and ego motion, while WoTE~\cite{li2025wote} predicts future BEV states for trajectory evaluation.
World4Drive~\cite{zheng2025world4drive} introduces spatial-semantic priors, and recent methods increasingly couple latent world modeling with planning, reasoning, and structured latent representations~\cite{xia2026drivelaw,liu2026driveworld,hong2026drivefuture,tan2025lcdrive,
wang2026latentwam,zhu2026dlwm,zhang2026deepsight,luo2026lastvla}.

Despite this progress, the relational semantics a latent world state should preserve often remain implicit.
Predicting future latent states encourages temporal scene modeling, while geometric priors and planning objectives provide spatial and task-level supervision, but do not explicitly constrain relationships between traffic participants.
For driving, such relationships are particularly important, since an actor's relevance depends not only on its appearance but also on its relative position and interaction with the ego vehicle~\cite{wild2026bridging}.
GraphWorld~\cite{song2026graphworld} explicitly incorporates an ego-centric interaction graph into its latent world representation and planning architecture, while GraphPilot~\cite{schmidt2025graphpilot} shows that structured scene-graph supervision benefits language-based driving.
However, these approaches either embed relational structure in the model or target it for language-based reasoning.
Whether relational scene structure can instead serve as \emph{privileged semantic supervision} for predictive world representations without requiring scene graphs at inference remains largely unexplored.

\begin{figure}[t]
    \centering
    \includegraphics[width=\columnwidth]{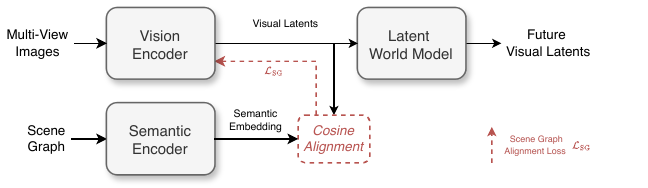}
    \caption{
        Overview of the proposed training-only relational grounding.
        Scene-graph semantics are aligned with visual latents through an additional
        cosine-alignment loss to encourage interaction-relevant relational structure
        in the learned representation, while the world-model inference pathway
        remains unchanged.
    }
    \label{fig:overview}
    \vspace{-0.5em}
\end{figure}

In this work, we investigate this question by relationally grounding the latent state of LAW using traffic scene graphs available only during training.
We serialize actor-centric graphs constructed from nuScenes~\cite{caesar2020nuscenes} 3D annotations into structured relational text and embed them using a frozen text encoder.
A lightweight alignment head maps LAW's visual latents into the same semantic space and aligns them with the scene-graph representation, as shown in Fig.~\ref{fig:overview}.
After training, the complete supervision pathway is discarded, requiring neither scene graphs, 3D annotations, nor additional computation at inference.
The objective is not specific to LAW and, in principle, applies to other latent driving architectures that expose a trainable scene representation.
We additionally study an optional extension that grounds LAW's predicted future latent state.

On nuScenes, structured scene-graph supervision reduces average trajectory L2 error from $0.661$ to $0.622$ and average collision rate from $0.456$ to $0.217$ relative to our controlled LAW retraining.
Compared with an unstructured caption-style semantic target, it further reduces L2 error by $7.2\%$ and collision rate by $30.4\%$.
Our work introduces training-only relational grounding for latent driving world models without modifying inference, shows the benefit of structured relational supervision over generic textual semantics, and analyzes how grounding affects observed and predicted future latent states.

\begin{figure*}[t]
    \centering
    \includegraphics[width=\textwidth]{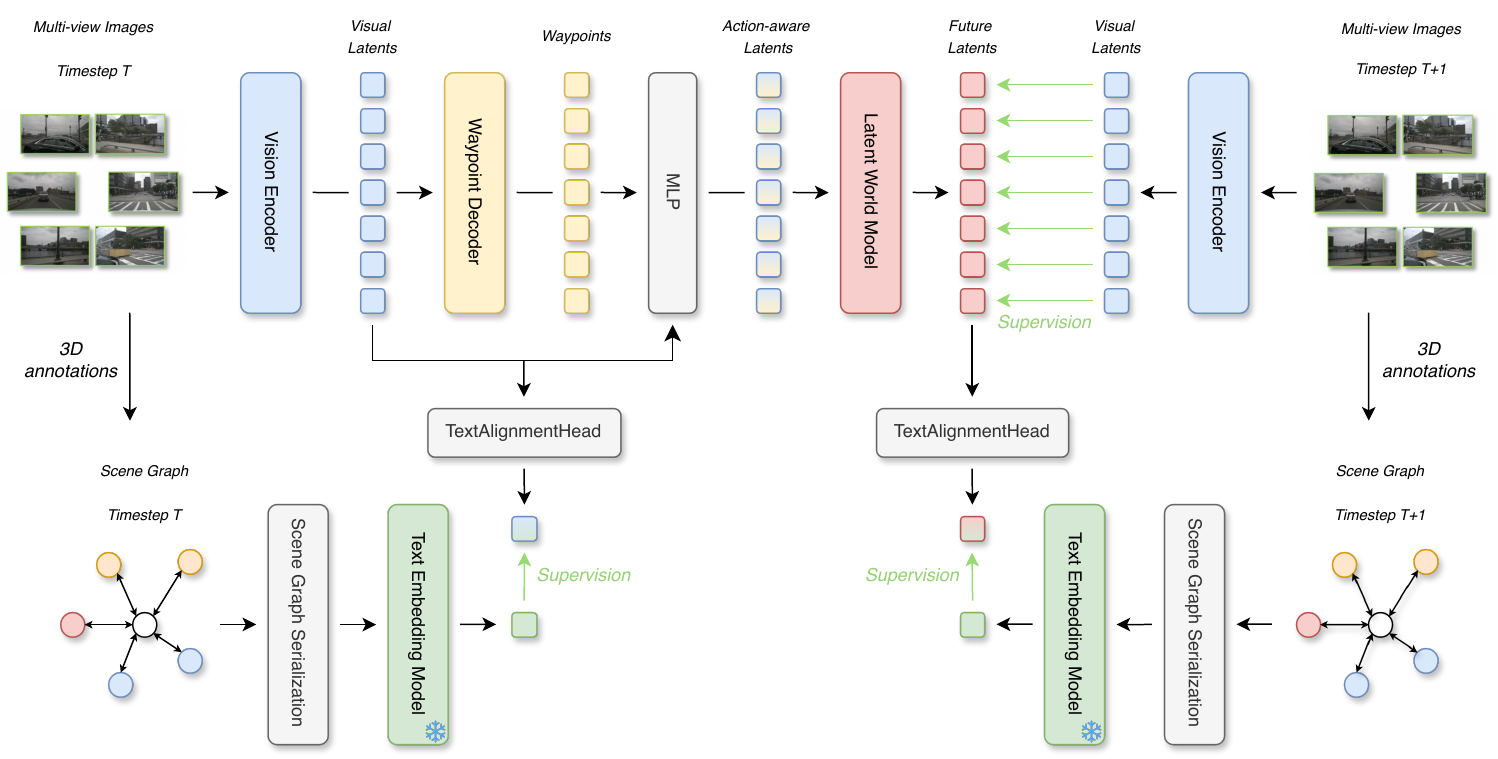}
    \caption{
        Overview of our relationally grounded latent world model.
        The upper branch shows the original LAW \cite{li2025enhancing} pipeline, which extracts visual latents from multi-view images, predicts ego waypoints, constructs action-aware latents, and predicts future visual latents.
        During training, we construct traffic scene graphs from the corresponding 3D annotations, serialize them, and encode them using a frozen text embedding model.
        We pool the visual latents at time $t$, project them into the semantic embedding space, and align them with the scene-graph embedding via cosine similarity.
        The right-hand auxiliary branch shows our optional temporal extension, which analogously aligns the predicted future latents with the semantic representation of the scene graph at $t+1$.
        The scene-graph, text-embedding, and alignment components are discarded after training and introduce no additional inference-time cost.
    }
    \label{fig:method}

\end{figure*}

\section{Relationally Grounded Latent World Model}
\label{sec:method}

We investigate whether structured relational scene information can serve as privileged semantic supervision for the latent state of a driving world model.
Building on LAW~\cite{li2025enhancing}, we introduce an auxiliary objective to align visual scene representations with the semantic embeddings of traffic scene graphs.
As illustrated in Fig.~\ref{fig:method}, we construct scene graphs from the 3D annotations for each training frame and use them only for semantic supervision.
We remove all additional components at inference.
We further investigate an optional temporal extension that also aligns the world model's predicted future representation with the scene graph at time $t+1$.

\subsection{Latent World Model}
\label{sec:law}

Given multi-view images $\mathbf{I}_t$ at time $t$, LAW~\cite{li2025enhancing} uses a vision encoder $E_{\mathrm{vis}}$ to extract a set of visual latents, $\mathbf{V}_t=E_{\mathrm{vis}}(\mathbf{I}_t)=\{\mathbf{v}_t^1,\ldots,\mathbf{v}_t^L\}$.
A waypoint decoder $D_{\mathrm{wp}}$ then predicts the future ego trajectory $\mathbf{W}_t=D_{\mathrm{wp}}(\mathbf{V}_t)$ from these latents.
To condition the world model on the planned ego motion, the waypoint sequence is flattened into a vector using $\operatorname{vec}(\cdot)$ and concatenated with each visual latent, yielding the action-aware features
\begin{equation}
    \mathbf{a}_t^i
    =
    \operatorname{MLP}
    \left(
        [\mathbf{v}_t^i,\operatorname{vec}(\mathbf{W}_t)]
    \right),
\end{equation}
where $\operatorname{vec}(\mathbf{W}_t)$ denotes the vectorized waypoint sequence obtained by concatenating all predicted waypoint coordinates.
The resulting latent set $\mathbf{A}_t=\{\mathbf{a}_t^i\}_{i=1}^{L}$ is processed by the latent world model $F_{\mathrm{WM}}$ to predict the future visual representation,
\begin{equation}
    \hat{\mathbf{V}}_{t+1}
    =
    F_{\mathrm{WM}}(\mathbf{A}_t),
\end{equation}
which is supervised by $\mathbf{V}_{t+1}=E_{\mathrm{vis}}(\mathbf{I}_{t+1})$ extracted from the corresponding future observation.
We denote LAW's original trajectory and latent-prediction objectives collectively as $\mathcal{L}_{\mathrm{LAW}}$.
Our approach leaves this architecture and its original supervision unchanged.

\subsection{Scene-Graph Semantic Supervision}
\label{sec:scene_graph_alignment}

For each training frame, we construct a relational traffic scene graph $\mathcal{G}_t=(\mathcal{N}_t,\mathcal{E}_t)$ from the corresponding nuScenes 3D annotations, where $\mathcal{N}_t$ denotes the set of graph nodes and $\mathcal{E}_t$ the set of pairwise relations.
We adopt the Actor-Only representation of GraphPilot~\cite{schmidt2025graphpilot}, in which the ego vehicle and surrounding traffic participants, such as vehicles, cyclists, and pedestrians, serve as nodes, while edges encode relations based on their relative orientation and proximity.
This provides a compact representation of the relational structure of the current traffic scene.

\begin{figure}[t]
    \centering
    \includegraphics[width=\columnwidth]{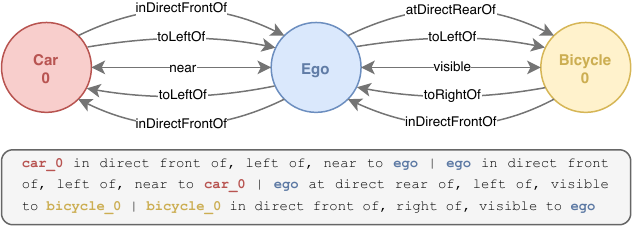}
    \caption{
        Example of an Actor-Only traffic scene graph and textual serialization, reproduced from GraphPilot~\cite{schmidt2025graphpilot}.
        Nodes represent traffic participants, while edges encode spatial, proximity, and visibility relations.
    }
    \label{fig:scene_graph_example}

    \vspace{-0.5em}
\end{figure}

Following GraphPilot~\cite{schmidt2025graphpilot}, we serialize the graph into
structured subject--relation--object statements, as illustrated in
Fig.~\ref{fig:scene_graph_example}.
We denote the resulting textual representation as
$s_t=\operatorname{Serialize}(\mathcal{G}_t)$.
To obtain a fixed semantic target, we encode $s_t$ using the frozen
Qwen3-Embedding model~\cite{zhang2025qwen3},
\begin{equation}
    \mathbf{z}_t^{\mathrm{SG}}
    =
    E_{\mathrm{text}}(s_t),
\end{equation}
where $E_{\mathrm{text}}$ denotes the text embedding model.

In parallel, the visual latents $\mathbf{V}_t$ are mapped into the same
embedding space using a learnable alignment head $P$.
The alignment head first performs mean pooling over the latent vectors to
obtain a scene-level representation and subsequently applies an MLP projection,
\begin{equation}
    \mathbf{z}_t^{\mathrm{vis}}
    =
    P(\mathbf{V}_t)
    =
    \operatorname{MLP}
    \left(
        \operatorname{MeanPool}(\mathbf{V}_t)
    \right).
\end{equation}
We align the resulting visual and scene-graph embeddings using cosine distance,
\begin{equation}
    \mathcal{L}_{\mathrm{SG}}
    =
    d_{\cos}
    \left(
        \mathbf{z}_t^{\mathrm{vis}},
        \mathbf{z}_t^{\mathrm{SG}}
    \right)
    =
    1 -
    \frac{
        \mathbf{z}_t^{\mathrm{vis}}
        \cdot
        \mathbf{z}_t^{\mathrm{SG}}
    }{
        \left\lVert \mathbf{z}_t^{\mathrm{vis}} \right\rVert_2
        \left\lVert \mathbf{z}_t^{\mathrm{SG}} \right\rVert_2
    }.
    \label{eq:scene_graph_loss}
\end{equation}
This auxiliary objective encourages the visual representation to capture the
high-level relational semantics encoded by the scene graph.
Gradients propagate through the visual encoder and alignment head, whereas the
text embedding model remains frozen.

\subsection{Temporal Semantic Alignment}
\label{sec:temporal_alignment}

As illustrated by the right-hand alignment branch in Fig.~\ref{fig:method}, we construct the future scene graph $\mathcal{G}_{t+1}$ from the corresponding 3D annotations and encode its serialized representation using the same frozen text encoder,
\begin{equation}
    \mathbf{z}_{t+1}^{\mathrm{SG}}
    =
    E_{\mathrm{text}}
    \left(
        \operatorname{Serialize}(\mathcal{G}_{t+1})
    \right).
\end{equation}
In parallel, the predicted future latents $\hat{\mathbf{V}}_{t+1}$ are mapped into the same semantic embedding space using the alignment head $P$ introduced above,
\begin{equation}
    \hat{\mathbf{z}}_{t+1}^{\mathrm{vis}}
    =
    P(\hat{\mathbf{V}}_{t+1})
    =
    \operatorname{MLP}
    \left(
        \operatorname{MeanPool}(\hat{\mathbf{V}}_{t+1})
    \right).
\end{equation}
The temporal semantic alignment loss is defined analogously to
Eq.~\ref{eq:scene_graph_loss} as
\begin{equation}
    \mathcal{L}_{\mathrm{temp}}
    =
    d_{\cos}
    \left(
        \hat{\mathbf{z}}_{t+1}^{\mathrm{vis}},
        \mathbf{z}_{t+1}^{\mathrm{SG}}
    \right),
    \label{eq:temporal_loss}
\end{equation}
where $d_{\cos}(\cdot,\cdot)$ denotes the cosine distance defined in
Eq.~\ref{eq:scene_graph_loss}.
While LAW supervises $\hat{\mathbf{V}}_{t+1}$ using visual latents extracted from the future observation, $\mathcal{L}_{\mathrm{temp}}$ additionally encourages the predicted state to preserve the relational semantics of the corresponding future traffic scene.

\subsection{Training Objective and Inference}
The complete training objective combines the original LAW objective with the current-state and optional temporal alignment losses,
\begin{equation}
    \mathcal{L}
    =
    \mathcal{L}_{\mathrm{LAW}}
    +
    \lambda_{\mathrm{SG}}\mathcal{L}_{\mathrm{SG}}
    +
    \lambda_{\mathrm{temp}}\mathcal{L}_{\mathrm{temp}},
    \label{eq:total_loss}
\end{equation}
where $\lambda_{\mathrm{SG}}$ and $\lambda_{\mathrm{temp}}$ control the strength of current-state and future-state semantic supervision, respectively.
All scene-graph, text-embedding, and alignment components are used only during training and discarded at inference, leaving the original LAW inference pipeline unchanged.

\section{Experimental Evaluation}
\label{sec:experiments}

We evaluate whether structured scene-graph supervision improves LAW and analyze the effects of relational structure, alignment strength, and temporal semantic grounding on downstream planning performance.

\subsection{Experimental Setup}
\label{sec:experimental_setup}

We evaluate our approach on the nuScenes~\cite{caesar2020nuscenes} planning benchmark using the perception-free LAW architecture~\cite{li2025enhancing}.
While LAW itself operates directly on multi-view images, our method additionally uses the available nuScenes 3D annotations during training to construct the scene-graph supervision targets described in Sec.~\ref{sec:scene_graph_alignment}.
We use these annotations only as privileged training supervision and do not require them at inference. 
Following LAW, we report the L2 displacement error of the predicted ego trajectory and the collision rate at prediction horizons of 1, 2, and 3 seconds, together with their averages.
Lower values are better for both metrics.

For all experiments, we train LAW from scratch using the official hyperparameters and training schedule on four NVIDIA H200 GPUs.
Apart from the additional semantic supervision objective, all variants use the same LAW architecture and training setup.
To ensure a controlled comparison, we retrain the LAW baseline under the same setup and report all improvements relative to this model rather than the published results or released checkpoint.
Unless stated otherwise, we use the structured scene-graph serialization from Sec.~\ref{sec:scene_graph_alignment}, a frozen Qwen3-Embedding-8B model, $\lambda_{\mathrm{SG}}=0.1$, and we set $\lambda_{\mathrm{temp}}=0$ and use
temporal alignment only in the corresponding ablation.
We use all scene-graph and semantic-alignment components only during training and remove them at inference.

\subsection{Structured Relational Supervision}
\label{sec:main_results}

We first evaluate whether aligning the visual latent state with a semantic scene-graph embedding improves downstream planning.
To distinguish structured relational supervision from generic textual semantic alignment, we additionally train an unstructured-text variant using the same frozen embedding model and alignment objective.
Instead of explicit subject--relation--object statements, this baseline describes the scene in caption-style prose, providing the same semantic information without the explicit graph structure and relation-centric representation used by our method.

As shown in Tab.~\ref{tab:main_results}, structured scene-graph supervision improves performance consistently across all prediction horizons.
Compared with our retrained LAW baseline, the average L2 error decreases from $0.661$ to $0.622$, corresponding to a relative reduction of $5.9\%$.
The improvement in collision rate is considerably larger, decreasing from $0.456$ to $0.217$, a relative reduction of $52.4\%$.

The comparison with unstructured text further suggests that this gain is not explained solely by introducing an auxiliary semantic target.
Although unstructured semantic alignment reduces the average collision rate to $0.312$, it slightly increases the average L2 error from $0.661$ to $0.670$.
In contrast, structured scene-graph supervision improves both metrics.
Relative to the unstructured representation, it reduces the average L2 error by $7.2\%$ and the average collision rate by $30.4\%$.
These results suggest that the structured scene-graph representation provides a more effective semantic target for the visual latent state than the unstructured textual baseline.

We hypothesize that this advantage arises from the actor-centric relational inductive bias of the scene graph.
Whereas caption-style text describes scene content in a less constrained form, the scene graph explicitly emphasizes relative orientation and proximity between traffic participants, which are directly relevant to interaction-aware driving.
The resulting target may therefore encourage LAW's visual latents to preserve relational information that its original trajectory and future-latent objectives only indirectly constrain.
This interpretation is also consistent with the particularly large reduction in collision rate, since collision avoidance depends strongly on representing the spatial relationships between the ego vehicle and surrounding actors.

\begin{table}[t!]
    \centering
    \caption{
        Influence of structured scene-graph supervision on nuScenes.
    }
    \label{tab:main_results}
    \resizebox{\columnwidth}{!}{
    \begin{tabular}{lcccccccc}
        \toprule
        & \multicolumn{4}{c}{L2 [m] $\downarrow$}
        & \multicolumn{4}{c}{Collision [\%] $\downarrow$} \\
        \cmidrule(lr){2-5}
        \cmidrule(lr){6-9}
        Method
        & 1s & 2s & 3s & Avg.
        & 1s & 2s & 3s & Avg. \\
        \midrule
        LAW
        & 0.327 & 0.628 & 1.029 & 0.661
        & 0.156 & 0.278 & 0.934 & 0.456 \\

        \midrule

        Unstructured Text
        & 0.310 & 0.632 & 1.067 & 0.670
        & 0.166 & 0.229 & 0.540 & 0.312 \\

        \rowcolor{gray!15}
        \textbf{Scene Graph}
        & \textbf{0.287} & \textbf{0.585} & \textbf{0.994} & \textbf{0.622}
        & \textbf{0.098} & \textbf{0.137} & \textbf{0.417} & \textbf{0.217} \\
        \bottomrule
    \end{tabular}
    }

\end{table}

\begin{table}[t!]
    \centering
    \caption{
        Ablation of current-state and temporal semantic alignment.
    }
    \label{tab:alignment_ablation}
    \begin{tabular}{lcc|cc}
        \toprule
        & $\lambda_{\mathrm{SG}}$
        & $\lambda_{\mathrm{temp}}$
        & Avg. L2 [m] $\downarrow$
        & Avg. Collision [\%] $\downarrow$ \\
        \midrule

        \multirow{6}{*}{\rotatebox{90}{\textit{Current-state}}} 
        & 0.00 & 0.00 & 0.661 & 0.456 \\
        & 0.05 & 0.00 & 0.660 & 0.283 \\
        & \cellcolor{gray!15}\textbf{0.10} & \cellcolor{gray!15}\textbf{0.00} & \cellcolor{gray!15}0.622 & \cellcolor{gray!15}\textbf{0.217} \\
        & 0.25 & 0.00 & 0.696 & 0.342 \\
        & 0.50 & 0.00 & 0.642 & 0.258 \\
        & 1.00 & 0.00 & 0.642 & 0.312 \\

        \midrule
        \multirow{6}{*}{\rotatebox{90}{\textit{Temporal}}} 
        & 0.00 & 0.10 & 0.628 & 0.476 \\
        & 0.00 & 0.50 & 0.651 & 0.659 \\
        & 0.05 & 0.05 & 0.632 & 0.257 \\
        & 0.10 & 0.10 & 0.624 & 0.291 \\
        & 0.25 & 0.25 & 0.606 & 0.436 \\
        & 0.50 & 0.50 & \textbf{0.594} & 0.387 \\

        \bottomrule
    \end{tabular}

\end{table}

\subsection{Semantic Alignment Ablations}
\label{sec:alignment_ablations}

We further analyze the influence of the semantic alignment objectives by varying the current-state loss weight $\lambda_{\mathrm{SG}}$ and evaluating the optional temporal alignment introduced in Sec.~\ref{sec:temporal_alignment}.
Tab.~\ref{tab:alignment_ablation} presents the results.

\paragraph{Alignment strength.}
The upper part of Tab.~\ref{tab:alignment_ablation} shows that the effect of current-state semantic supervision is not monotonic in $\lambda_{\mathrm{SG}}$.
A small weight of $0.05$ already reduces the average collision rate from $0.456$ to $0.283$, while leaving the trajectory error almost unchanged.
Increasing the weight to $0.1$ yields the best overall trade-off, reducing the average L2 error to $0.622$ and the collision rate to $0.217$.
Larger weights do not provide consistent additional gains, suggesting that scene-graph alignment is most effective as an auxiliary constraint rather than a dominant training objective.

\paragraph{Temporal alignment.}
The lower part of Tab.~\ref{tab:alignment_ablation} evaluates semantic supervision of the predicted future state.
Future-state alignment alone improves trajectory error only at moderate weight: $\lambda_{\mathrm{temp}}=0.1$ reduces average L2 from $0.661$ to $0.628$, but does not improve collision performance.
Increasing the future-only weight to $0.5$ largely removes the L2 gain and substantially worsens collision performance, indicating that overly strong supervision of the predicted latent state can be detrimental.

Joint current--future supervision exhibits a different trade-off.
Increasing both alignment weights progressively improves trajectory accuracy, with $\lambda_{\mathrm{SG}}=\lambda_{\mathrm{temp}}=0.5$ achieving the lowest average L2 error of $0.594$, a $10.1\%$ reduction over the baseline.
However, its collision rate of $0.387$ remains substantially above the $0.217$ obtained with current-state alignment alone.
Overall, these results suggest that relational grounding is most effective for collision reduction when applied to the current visual state, whereas temporal grounding can further improve trajectory accuracy but requires careful weighting and does not provide the same safety benefit.

\section{Conclusion}
\label{sec:conclusion}

We investigated whether relational scene structure can serve as privileged semantic supervision for predictive latent world representations in autonomous driving.
By aligning LAW's visual latents with semantic embeddings of actor-centric traffic scene graphs during training, we improve both trajectory accuracy and collision performance without requiring scene graphs, 3D annotations, or additional inference computation.
Our experiments further show that structured scene-graph supervision outperforms an unstructured caption-style semantic target, supporting the hypothesis that explicitly encoding relations among traffic participants provides a useful inductive bias for learning latent world-model representations.
Temporal grounding of predicted future states can further improve trajectory accuracy, but it exhibits a different trade-off with collision performance.

Our current formulation obtains the semantic target by serializing the scene graph and encoding it with a frozen text model, making the target potentially sensitive to serialization choices.
Future work will therefore investigate graph-native encoders, such as relational GNNs or graph transformers, that directly exploit graph topology, learn task-relevant entity and relation weighting, and provide compact relational supervision without requiring textual serialization.
We further plan to study the transfer of this training-only objective to other latent driving world models beyond LAW.


\bibliographystyle{IEEEtran}
\bibliography{bibliography}

@inproceedings{shao2024lmdrive,
  title={Lmdrive: Closed-loop end-to-end driving with large language models},
  author={Shao, Hao and Hu, Yuxuan and Wang, Letian and Song, Guanglu and Waslander, Steven L and Liu, Yu and Li, Hongsheng},
  booktitle={Proceedings of the IEEE/CVF Conference on Computer Vision and Pattern Recognition},
  pages={15120--15130},
  year={2024}
}

@inproceedings{renz2025simlingo,
    title={Simlingo: Vision-only closed-loop autonomous driving with language-action alignment},
    author={Renz, Katrin and Chen, Long and Arani, Elahe and Sinavski, Oleg},
    booktitle={Proceedings of the Computer Vision and Pattern Recognition Conference},
    pages={11993--12003},
    year={2025}
}

@inproceedings{hu2023planning,
  title={Planning-oriented autonomous driving},
  author={Hu, Yihan and Yang, Jiazhi and Chen, Li and Li, Keyu and Sima, Chonghao and Zhu, Xizhou and Chai, Siqi and Du, Senyao and Lin, Tianwei and Wang, Wenhai and others},
  booktitle={Proceedings of the IEEE/CVF conference on computer vision and pattern recognition},
  pages={17853--17862},
  year={2023}
}

@article{chen2024vadv2,
  title={Vadv2: End-to-end vectorized autonomous driving via probabilistic planning},
  author={Chen, Shaoyu and Jiang, Bo and Gao, Hao and Liao, Bencheng and others},
  journal={arXiv preprint arXiv:2402.13243},
  year={2024}
}

@article{schmidt2025graphpilot,
  title={GraphPilot: Grounded Scene Graph Conditioning for Language-Based Autonomous Driving},
  author={Schmidt, Fabian and Enzweiler, Markus and Valada, Abhinav},
  journal={arXiv preprint arXiv:2511.11266},
  year={2025}
}

@inproceedings{li2025enhancing,
  title={Enhancing end-to-end autonomous driving with latent world model},
  author={Li, Yingyan and Fan, Lue and He, Jiawei and Wang, Yuqi and Chen, Yuntao and Zhang, Zhaoxiang and Tan, Tieniu},
  booktitle={International Conference on Learning Representations},
  volume={2025},
  pages={42942--42959},
  year={2025}
}

@article{zhang2025qwen3,
  title={Qwen3 embedding: Advancing text embedding and reranking through foundation models},
  author={Zhang, Yanzhao and Li, Mingxin and Long, Dingkun and Zhang, Xin and Lin, Huan and Yang, Baosong and Xie, Pengjun and Yang, An and Liu, Dayiheng and Lin, Junyang and others},
  journal={arXiv preprint arXiv:2506.05176},
  year={2025}
}

@inproceedings{caesar2020nuscenes,
  title={nuscenes: A multimodal dataset for autonomous driving},
  author={Caesar, Holger and Bankiti, Varun and Lang, Alex H and Vora, Sourabh and Liong, Venice Erin and Xu, Qiang and Krishnan, Anush and Pan, Yu and Baldan, Giancarlo and Beijbom, Oscar},
  booktitle={2020 IEEE/CVF conference on computer vision and pattern recognition (CVPR)},
  pages={11618--11628},
  year={2020},
  organization={IEEE}
}

@inproceedings{li2025wote,
  title={End-to-end driving with online trajectory evaluation via bev world model},
  author={Li, Yingyan and Wang, Yuqi and Liu, Yang and He, Jiawei and Fan, Lue and Zhang, Zhaoxiang},
  booktitle={2025 IEEE/CVF International Conference on Computer Vision (ICCV)},
  pages={27137--27146},
  year={2025},
  organization={IEEE}
}

@inproceedings{zheng2025world4drive,
  title={World4drive: End-to-end autonomous driving via intention-aware physical latent world model},
  author={Zheng, Yupeng and Yang, Pengxuan and Xing, Zebin and Zhang, Qichao and Zheng, Yuhang and Gao, Yinfeng and Li, Pengfei and Zhang, Teng and Xia, Zhongpu and Jia, Peng and others},
  booktitle={2025 IEEE/CVF International Conference on Computer Vision (ICCV)},
  pages={28632--28642},
  year={2025},
  organization={IEEE}
}

@inproceedings{xia2026drivelaw,
  title={Drivelaw: Unifying planning and video generation in a latent driving world},
  author={Xia, Tianze and Li, Yongkang and Zhou, Lijun and Yao, Jingfeng and Xiong, Kaixin and Sun, Haiyang and Wang, Bing and Ma, Kun and Chen, Guang and Ye, Hangjun and others},
  booktitle={Proceedings of the IEEE/CVF Conference on Computer Vision and Pattern Recognition},
  pages={39701--39712},
  year={2026}
}

@article{liu2026driveworld,
  title={Driveworld-vla: Unified latent-space world modeling with vision-language-action for autonomous driving},
  author={Liu, Lin and Song, Ziying and Jia, Caiyan and Ye, Hangjun and Hao, Xiaoshuai and Chen, Long and others},
  journal={arXiv preprint arXiv:2602.06521},
  year={2026}
}

@article{hong2026drivefuture,
  title={DriveFuture: Future-Aware Latent World Models for Autonomous Driving},
  author={Hong, Yufeng and Zhou, Xiaotian and Li, Yingyan and Zhou, Xiangpo and Liu, Lin and Luo, Yadan and Xu, Shaoqing and Yang, Lei and Song, Ziying},
  journal={arXiv preprint arXiv:2605.09701},
  year={2026}
}

@article{wang2026latentwam,
  title={Latent-wam: Latent world action modeling for end-to-end autonomous driving},
  author={Wang, Linbo and Zheng, Yupeng and Chen, Qiang and Li, Shiwei and Zhang, Yichen and Xing, Zebin and Zhang, Qichao and Li, Xiang and Qian, Deheng and Yang, Pengxuan and others},
  journal={arXiv preprint arXiv:2603.24581},
  year={2026}
}

@article{zhu2026dlwm,
  title={DLWM: Dual Latent World Models enable Holistic Gaussian-centric Pre-training in Autonomous Driving},
  author={Zhu, Yiyao and Xue, Ying and Zhang, Haiming and Jiang, Guangfeng and Zhou, Wending and Yan, Xu and Gao, Jiantao and Cai, Yingjie and Liu, Bingbing and Li, Zhen and others},
  journal={arXiv preprint arXiv:2604.00969},
  year={2026}
}

@article{zhang2026deepsight,
  title={DeepSight: Long-Horizon World Modeling via Latent States Prediction for End-to-End Autonomous Driving},
  author={Zhang, Lingjun and Wu, Changjie and Shi, Linzhe and Li, Jiangyang and Liu, Jiaxin and Yang, Lei and Zhang, Hang and Xu, Mu and Wang, Hong},
  journal={arXiv preprint arXiv:2605.10564},
  year={2026}
}

@article{song2026graphworld,
  title={GraphWorld: Long-Horizon Planning with World Models for End-to-End Autonomous Driving},
  author={Song, Ziying and Jia, Caiyan and Liu, Lin and Yang, Lei and Zhang, Shengkai and Jia, Feiyang and Zhao, Fengda and Wu, Peiliang and Xu, Shaoqing and Lv, Chen and others},
  journal={arXiv preprint arXiv:2606.16274},
  year={2026}
}

@article{tan2025lcdrive,
  title={Latent chain-of-thought world modeling for end-to-end driving},
  author={Tan, Shuhan and Chitta, Kashyap and Chen, Yuxiao and Tian, Ran and You, Yurong and Wang, Yan and Luo, Wenjie and Cao, Yulong and Krahenbuhl, Philipp and Pavone, Marco and others},
  journal={arXiv preprint arXiv:2512.10226},
  year={2025}
}

@article{luo2026lastvla,
  title={Last-vla: Thinking in latent spatio-temporal space for vision-language-action in autonomous driving},
  author={Luo, Yuechen and Li, Fang and Xu, Shaoqing and Ji, Yang and Zhang, Zehan and Wang, Bing and Shen, Yuannan and Cui, Jianwei and Chen, Long and Chen, Guang and others},
  journal={arXiv preprint arXiv:2603.01928},
  year={2026}
}

@inproceedings{zhou2026opendrivevla,
  title={Opendrivevla: Towards end-to-end autonomous driving with large vision language action model},
  author={Zhou, Xingcheng and Han, Xuyuan and Yang, Feng and Ma, Yunpu and Tresp, Volker and Knoll, Alois},
  booktitle={Proceedings of the AAAI Conference on Artificial Intelligence},
  volume={40},
  number={16},
  pages={13782--13790},
  year={2026}
}

@article{schmidt2026laddrive,
  title={Lad-drive: Bridging language and trajectory with action-aware diffusion transformers},
  author={Schmidt, Fabian and Fedurko, Karol and Enzweiler, Markus and Valada, Abhinav},
  journal={arXiv preprint arXiv:2603.02035},
  year={2026}
}

@inproceedings{sima2024drivelm,
  title={Drivelm: Driving with graph visual question answering},
  author={Sima, Chonghao and Renz, Katrin and Chitta, Kashyap and Chen, Li and Zhang, Hanxue and Xie, Chengen and Bei{\ss}wenger, Jens and Luo, Ping and Geiger, Andreas and Li, Hongyang},
  booktitle={European conference on computer vision},
  pages={256--274},
  year={2024},
  organization={Springer}
}

@article{gao2024vista,
  title={Vista: A generalizable driving world model with high fidelity and versatile controllability},
  author={Gao, Shenyuan and Yang, Jiazhi and Chen, Li and Chitta, Kashyap and Qiu, Yihang and Geiger, Andreas and Zhang, Jun and Li, Hongyang},
  journal={Advances in Neural Information Processing Systems},
  volume={37},
  pages={91560--91596},
  year={2024}
}

@inproceedings{wang2024drivedreamer,
  title={Drivedreamer: Towards real-world-drive world models for autonomous driving},
  author={Wang, Xiaofeng and Zhu, Zheng and Huang, Guan and Chen, Xinze and Zhu, Jiagang and Lu, Jiwen},
  booktitle={European conference on computer vision},
  pages={55--72},
  year={2024},
  organization={Springer}
}

@article{hu2023gaia,
  title={Gaia-1: A generative world model for autonomous driving},
  author={Hu, Anthony and Russell, Lloyd and Yeo, Hudson and Murez, Zak and Fedoseev, George and Kendall, Alex and Shotton, Jamie and Corrado, Gianluca},
  journal={arXiv preprint arXiv:2309.17080},
  year={2023}
}

@article{mousakhan2025orbis,
  title={Orbis: Overcoming challenges of long-horizon prediction in driving world models},
  author={Mousakhan, Arian and Mittal, Sudhanshu and Galesso, Silvio and Farid, Karim and Brox, Thomas},
  journal={arXiv preprint arXiv:2507.13162},
  year={2025}
}

@article{wild2026bridging,
  title={Bridging Structure and Language: Graph-Based Visual Reasoning for Autonomous Road Understanding},
  author={Wild, Lena and Luo, Katie Z and Pavone, Marco},
  journal={arXiv preprint arXiv:2605.20942},
  year={2026}
}

\addtolength{\textheight}{-12cm}   

\end{document}